\pdfoutput=1

\documentclass[11pt,hyphens]{article}

\usepackage{acl}

\usepackage{times}
\usepackage{latexsym}

\usepackage{times}  
\usepackage{helvet}  
\usepackage{courier}  
\usepackage{graphicx} 
\usepackage{natbib}  
\usepackage{caption} 
\usepackage{booktabs}
\usepackage{multirow}
\usepackage{subfig}
\usepackage{enumitem}

\usepackage[T1]{fontenc}

\usepackage[utf8]{inputenc}

\usepackage{microtype}

\usepackage{inconsolata}

\usepackage{graphicx}

\title{Enhancing Security and Strengthening Defenses in Automated Short-Answer Grading Systems}

\author{
Sahar Yarmohammadtoosky\textsuperscript{1} \quad Yiyun Zhou\textsuperscript{2} \quad Victoria Yaneva\textsuperscript{2} \\
Peter Baldwin\textsuperscript{2} \quad Saed Rezayi\textsuperscript{2} \quad Brian Clauser\textsuperscript{2} \quad Polina Harik\textsuperscript{2} \\
\textsuperscript{1}School of Data Science and Analytics, Kennesaw State University \\
\textsuperscript{2}National Board of Medical Examiners (NBME) \\
\texttt{yarmohamadishr@gmail.com} \\
\texttt{YYZhou@nbme.org, victoriayaneva@gmail.com, PBaldwin@nbme.org} \\
\texttt{SRezayidemne@nbme.org, beclauser@gmail.com, PHarik@nbme.org}
}

\begin{document}
\maketitle
\begin{abstract}
This study examines vulnerabilities in transformer-based automated short-answer grading systems used in medical education, with a focus on how these systems can be manipulated through adversarial gaming strategies. Our research identifies three main types of gaming strategies that exploit the system's weaknesses, potentially leading to false positives. To counteract these vulnerabilities, we implement several adversarial training methods designed to enhance the system’s robustness. Our results indicate that these methods significantly reduce the susceptibility of grading systems to such manipulations, especially when combined with ensemble techniques like majority voting and Ridge regression, which further improve the system's defense against sophisticated adversarial inputs. Additionally, employing large language models such as GPT-4 with varied prompting techniques has shown promise in recognizing and scoring gaming strategies effectively. The findings underscore the importance of continuous improvements in AI-driven educational tools to ensure their reliability and fairness in high-stakes settings.
\end{abstract}

\section{Introduction}
\label{sec:intro}
As technology advances, automated scoring of free-text responses is transforming how we evaluate written answers, making the process faster and more consistent \cite{yannakoudakis2011new}. Early research in this area has focused on instance-based methods, treating the task as a supervised text classification problem \cite{kumar2019get}. In this approach, models are trained using labeled data to predict labels for unseen data, such as predicting whether a short answer submitted to an Automated Short Answer Grading (ASAG) system is correct or incorrect \cite{bonthu2021automated}. More recently, some ASAG systems have taken a similarity-based approach, where each new response is assigned the label of the response it most closely matches from a sample of previously annotated responses. Neural similarity-based models have further advanced this field by learning rich response (or question-response) embeddings and matching them using cosine similarity, demonstrating superior performance in capturing meaning beyond surface-level text \cite{schneider2022towards}. 

Despite the significant potential demonstrated by similarity-based ASAG models, these models are especially vulnerable to scoring errors when presented with certain kinds of responses 
(Section \ref{sec:background}). This creates an opportunity for examinees to exploit these vulnerabilities to earn undeserved credit, which can erode trust in automated grading and raise concerns over the responsible use of AI in educational assessments. Deliberate attempts by examinees to exploit ASAG systems in this way are known as ``gaming strategies."

\begin{table*}[ht]
\centering
\footnotesize
\begin{tabular}{@{}p{3.3cm}p{13cm}@{}}
\toprule
\textbf{Component} & \textbf{Description} \\ \midrule
\textbf{Stem} & A previously healthy 26-year-old man is brought to the emergency department because of a tingling sensation in his fingers and toes for 3 days and progressive weakness of his legs. He had an upper respiratory tract infection 2 weeks ago. He has not traveled recently. He was unable to get up from bed this morning and called the ambulance. Temperature is 37.3\textdegree{}C (99.1\textdegree{}F), pulse is 110/min, respirations are 22/min, and blood pressure is 128/82 mm Hg. Pulse oximetry on room air shows an oxygen saturation of 99\%. Physical examination shows weakness of all four extremities in flexion and extension; this weakness is increased in the distal compared with the proximal muscle groups. Deep tendon reflexes are absent throughout. The sensation is mildly decreased over both feet. \\ \addlinespace
\textbf{Lead-in} & What is the most likely diagnosis? \\ \addlinespace
\textbf{Sample Correct Answers} & Guillain-Barré syndrome; acute immune-mediated polyneuropathy \\ \bottomrule
\end{tabular}
\caption{The parts of a short-answer question in the medical domain.}
\label{table:short-answer}
\end{table*}

The objective of this study is to identify and analyze various potential gaming strategies that students might use to manipulate or deceive ASAG systems. Furthermore, we propose several countermeasures designed to mitigate against these gaming attempts with a focus on addressing vulnerabilities in transformer-based and prompt-engineering methods for short-answer scoring systems within the context of medical education. The robustness of an ASAG system to gaming responses before and after these countermeasures were adopted is reported. These objectives were motivated by the following research questions:

\begin{enumerate}[noitemsep]
\item How vulnerable are transformer-based grading systems to adversarial gaming strategies commonly used by test takers when unsure of the correct answer?
\item What effect do the proposed adversarial training methods have on the robustness of these systems to such gaming strategies?
\item How effective are different prompt engineering strategies in improving the accuracy and reducing false positive rates (FPR) in automated scoring models when presented with adversarial inputs?
\end{enumerate}

This study advances the ASAG field by addressing the critical issue of vulnerability to adversarial gaming strategies. By identifying such strategies and developing effective countermeasures, the robustness, integrity, and reliability of transformer-based short-answer grading systems can be improved. The reported findings have broad practical benefits including improving the trustworthiness of automated grading tools in educational settings and contributing to the security of AI-driven systems against adversarial attacks. The technical advancements that are reported are also complemented by theoretical insights into the challenge posed by gaming in the context of ASAG specifically as well as into the responsible use of AI in education more generally.

\section{Related Work}
\label{sec:background}


With the advent of transformer models, neural similarity-based ASAG techniques have demonstrated improved accuracy and reduced data annotation requirements compared to instance-based methods \cite{bexte2023similarity}. However, these advancements have also introduced new challenges, particularly the susceptibility of similarity-based systems to adversarial attacks \citep{filighera2020fooling}. Such attacks can range from submitting random strings of letters \citep{ding2020don} to adding irrelevant yet carefully chosen words to otherwise valid responses \citep{filighera2023cheating}, with the goal of deceiving the model into misclassification. For example, \citet{ding2020don} found that a nonsensical string like "nswvtnvakgxpm" could be classified as a correct response by an ASAG system.

Within the medical domain,  \citet{baldwin2025vulnerability} have shown that several gaming strategies were successful in "deceiving" a similarity-based system. These strategies consisted of entering the following as responses to the short-answer questions: (1) random number of words selected at random from the stem\footnote{An item stem is the part of a test question that presents the problem or scenario to be answered or responded to, as shown in Table \ref{table:short-answer}.}, (2) random number of consecutive words selected at random from the stem, (3) random number of medical terms selected at random from the stem, (4) keywords selected from the stem by a content expert, and (5) a summary of the stem produced by GPT 3.5, as well as (6) listing multiple responses only one of which is correct. The results showed that the first five strategies lead to a success rate between 6\% to 16\%, while the last strategy led to a success rate of 57\%, underscoring the need for addressing these vulnerabilities.


While prior work defined the problem of gaming strategies and quantified their effects on transformer-based scoring systems, this study focuses on systematically evaluating multiple adversarial training techniques and ensemble strategies to enhance system resilience within the clinical ASAG domain. Additionally, we explore the role of LLMs, such as GPT-4, in detecting and mitigating adversarial manipulation.

\section{Methodology}

\subsection{Dataset}

The dataset comprises 71 short-answer questions (SAQs) with 36,735 responses from 24,235 examinees. An example of an SAQ is shown in Table 1. Responses were collected during the administration of a Medicine Clinical Science subject exam distributed to a large number of medical schools in the US and Canada for use as a summative, end-of-semester exam.

\subsection{ACTA System}
Experiments were undertaken using the ACTA system (Analysis of Clinical Text for Assessment; \citet{suen2023acta}),  a transformer-based ASAG system designed to classify short responses to medical questions as correct or incorrect. To achieve this, ACTA utilizes sentence BERT \citep{reimers2019sentence} and contrastive learning. When presented with a new response, ACTA matches it to the most similar response within a training set of human-scored responses and assigns it the matched response's label (correct or incorrect), provided their similarity exceeds a given operational threshold (for a detailed description of ACTA, see \citet{suen2023acta}). 
While ACTA achieves near human-level performance with a binary F1 score of .98, previously reported weaknesses of transformer-based grading systems require an investigation of ACTA’s susceptibility to gaming. 

\subsection{Gaming Strategies Method}

Following \citet{baldwin2025vulnerability}, we simulate three gaming strategies meant to resemble how students \textit{without} the requisite knowledge of a correct answer might nevertheless respond to an item. Data were generated as follows:

\begin{enumerate}[noitemsep]
  \item Simulate responses by randomly sampling words (excluding stop words) from a given item's clinical vignette. Variations of this strategy include consecutive words, non-consecutive words, and samples of words that appear in both the item description \textit{and} a generic list of medical terms.
  \item Utilize a summary of the clinical scenario as a response. Summaries were obtained using ChatGPT.
  \item Utilize ``mixed" responses that combine both correct and plausible incorrect answers into a single response, which, following operational guidelines, should be scored as incorrect.
\end{enumerate}
For our data, the strategies 
generated an impractically large number of responses. To create a set of responses that could feasibly be used as part of an operational process, we randomly sample 5\% from each strategy, resulting in 14,657, 573, and 584 simulated responses for strategies 1, 2, and 3, respectively. While simulated responses were largely nonce phrases or unequivocally incorrect, 3 simulated responses exactly matched (real) correct responses from the training data. Three misclassifications were deemed tolerable for our purpose, and all artificial responses were designated as incorrect.

\begin{figure*}[htbp!]
    \centering
    \begin{minipage}{0.45\textwidth} 
        \centering
        \includegraphics[width=1\columnwidth]{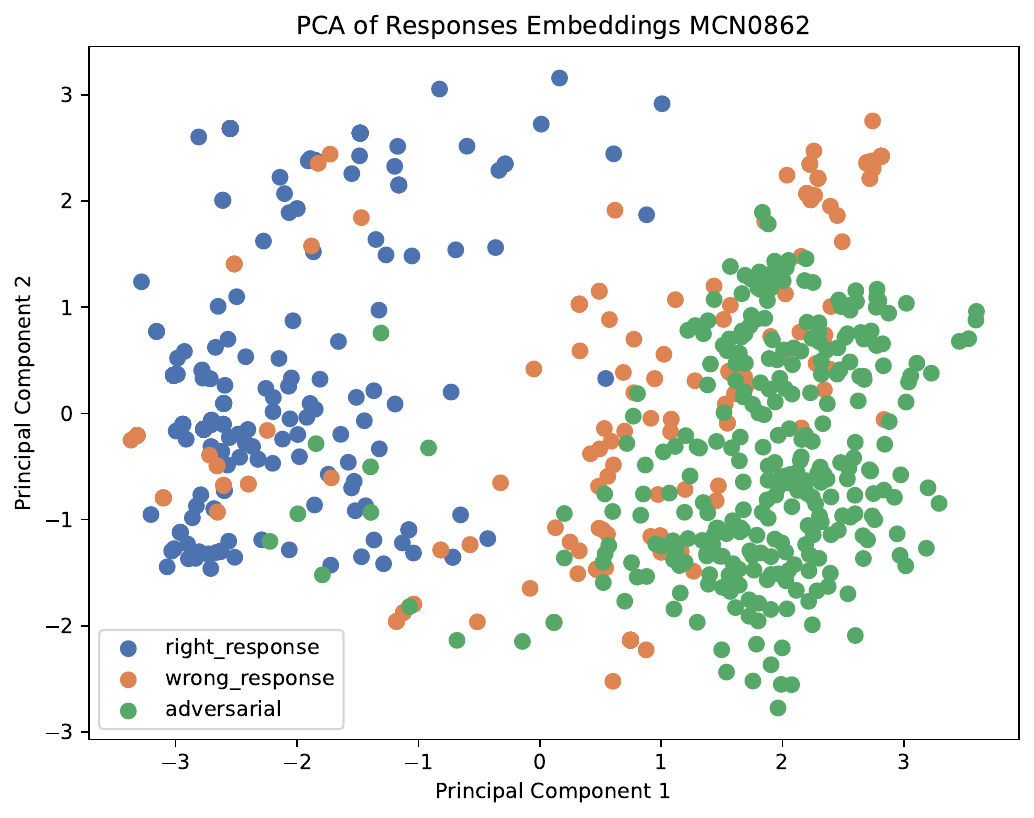} 
        \caption{PCA of Response Embedding for Item 1}
        \label{fig:figure1}
    \end{minipage}\hfill
    \begin{minipage}{0.45\textwidth} 
        \centering
        \includegraphics[width=1\columnwidth]{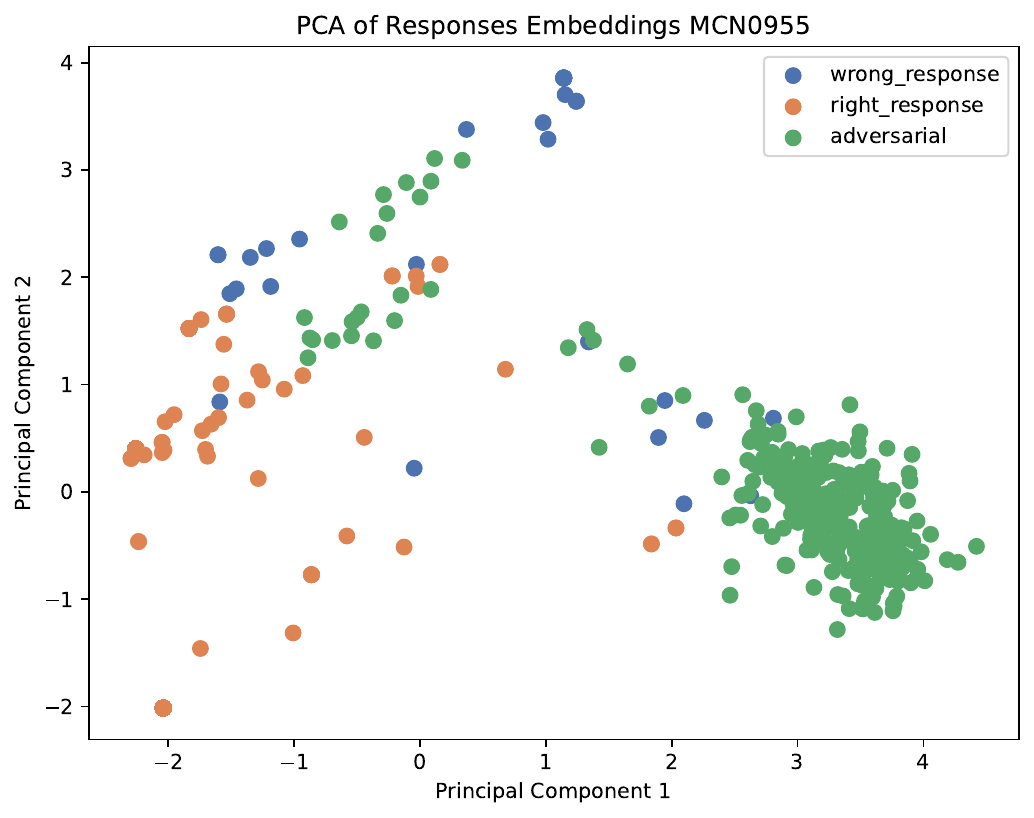} 
        \caption{PCA of Response Embedding for Item 2}
        \label{fig:figure2}
    \end{minipage}
\end{figure*}

Following a principal component analysis (PCA), Figures~\ref{fig:figure1} and \ref{fig:figure2} plot the responses for two SAQs in the space defined by principal components 1 and 2. Differences in the identification of adversarial examples across items can be observed. For SAQ~1 (Figure~\ref{fig:figure1}), the distribution of gaming responses shares considerable overlap with the distribution of correct responses, suggesting that gaming responses may have a relatively high probability of being misclassified. In contrast, for SAQ~2 (Figure~\ref{fig:figure2}), gaming responses are comparatively isolated, suggesting that these responses may be more readily identified by an ASAG system.

\subsection{Prompt Engineering Method}

Using large language models to score the real dataset has already shown promising results. This motivated the use of these models with different prompting techniques to evaluate whether large language models can accurately recognize and score gaming responses. Due to the consistently strong performance demonstrated by \textbf{ChatGPT4} \cite{achiam2023gpt} across various experimental settings, this model was selected as the primary tool for conducting this series of experiments.

\section{Experiments and Results} 

This section presents the initial evaluation of the ACTA system’s vulnerability to gaming before and after adversarial training as well as the results from the prompt engineering experiments.

\begin{figure*}[ht]
\centering
\includegraphics[scale=0.4]{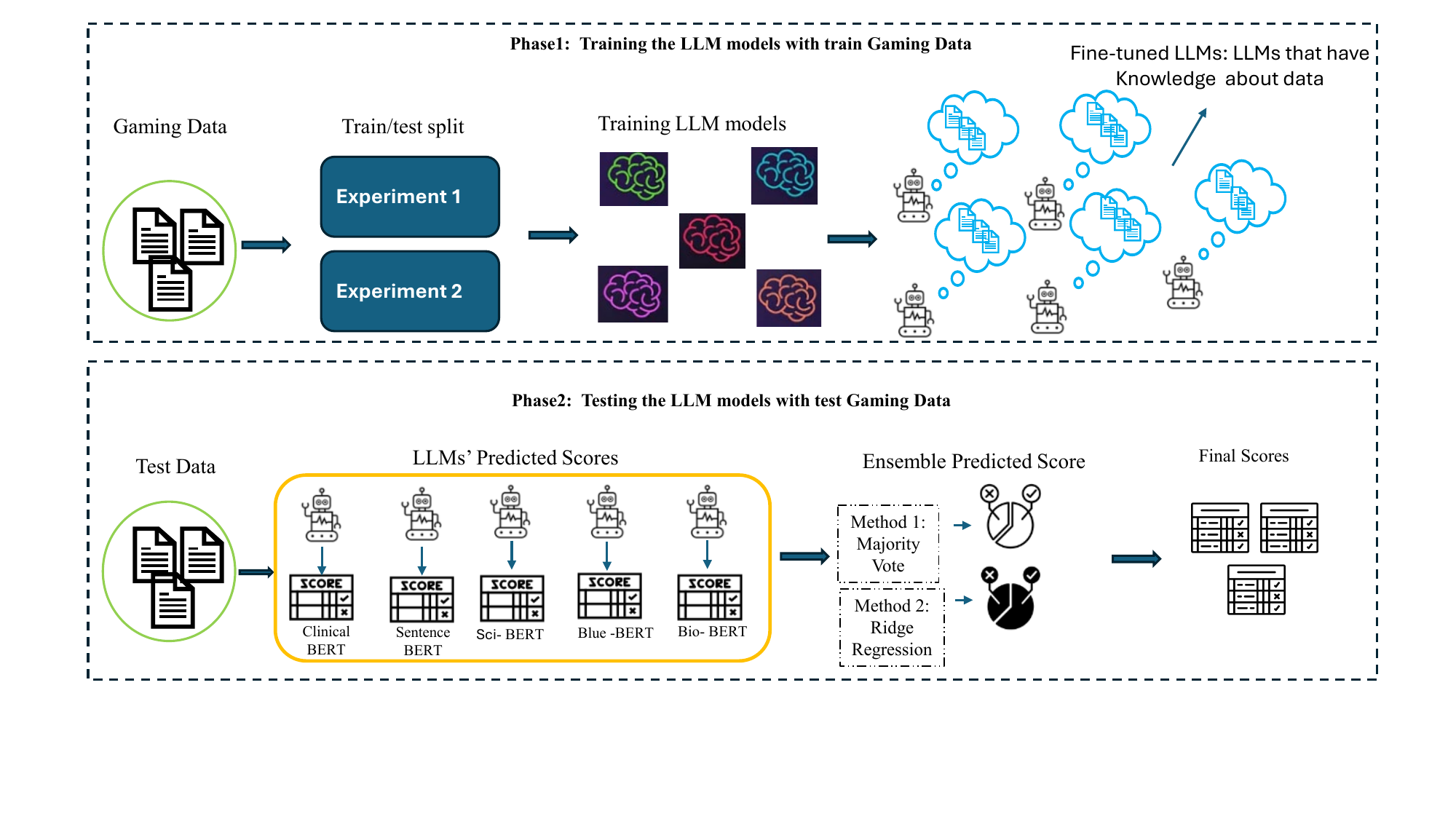}  
\vspace{-30pt}  
\caption{Adversarial Defense Workflow. Gaming data is combined with real data for training and testing purposes.}
\label{fig_workflow}
\end{figure*}

\subsection{ACTA Performance with Gaming Data}

We began by evaluating ACTA’s scoring of gaming responses prior to any adversarial training. The model was trained on 70\% (26,095) of the real responses and evaluated on the remaining 30\% (10, 890) combined with all artificial responses. Since the number of simulated gaming responses varies across strategies and experiments, we report two separate measures: F1 for real responses and false positive rate (FPR) for artificial responses. ACTA performed well when scoring real data (F1 = .9845); however, the gaming strategies deceived ACTA into misclassifying many of the artificial responses as ``correct." FPRs for strategies 1, 2, and 3 were .061, .189, and .435, respectively, demonstrating the vulnerability of this system to examinee gaming (Table \ref{tab:detector_table}). Responses from strategy 3 were especially challenging to classify correctly, illustrating the potential for examinees with partial knowledge to game systems that have not been adversarial trained by simply listing as many plausible answers as possible.

\subsection{Adversarial Training Experiments}

\begin{figure*}[htbp!]
    \centering
    \begin{minipage}{0.45\textwidth} 
        \centering
        \includegraphics[width=1\columnwidth]{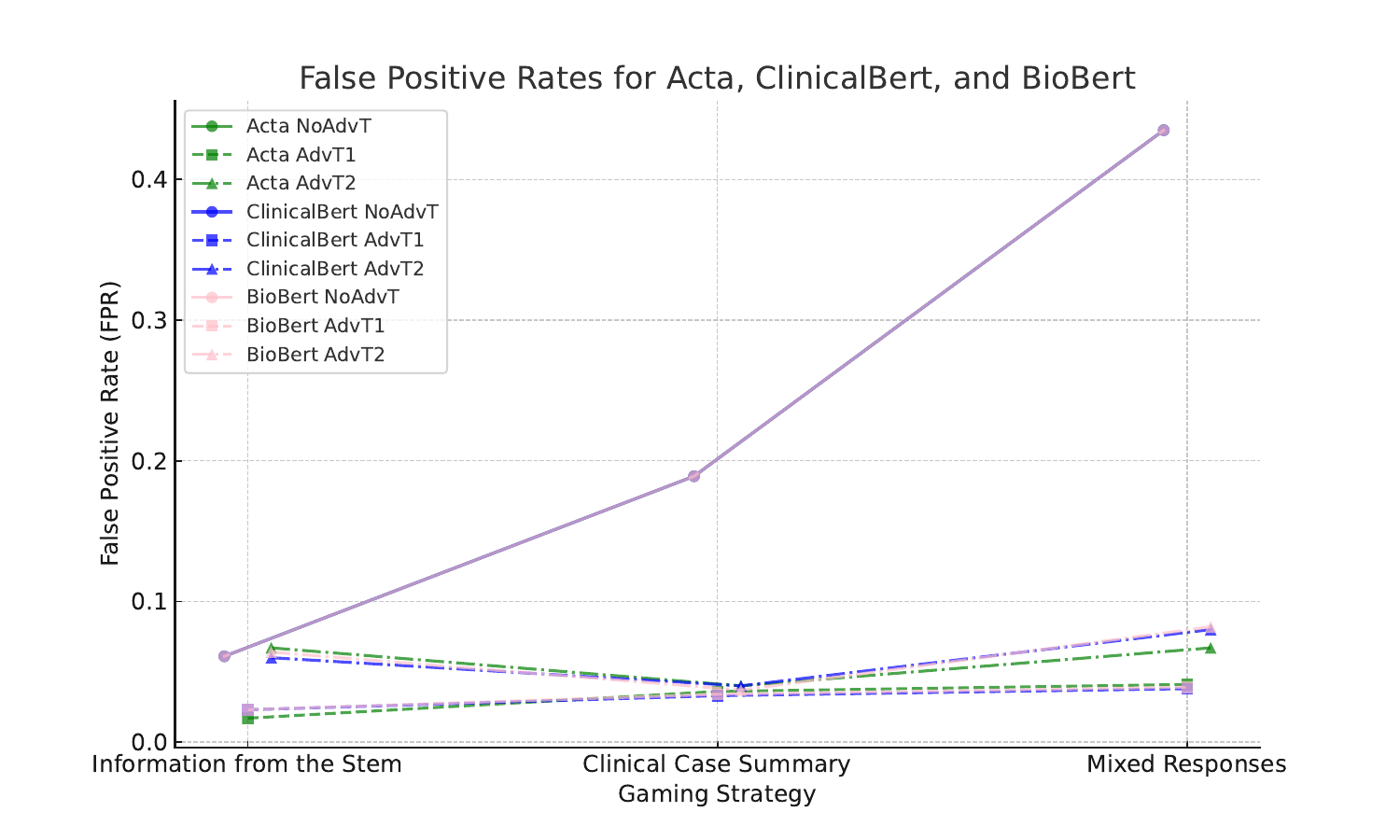} 
        \caption{FPR Across Different Models - Part 1}
        \label{fig:fpr_majority}
    \end{minipage}\hfill
    \begin{minipage}{0.45\textwidth} 
        \centering
        \includegraphics[width=1\columnwidth]{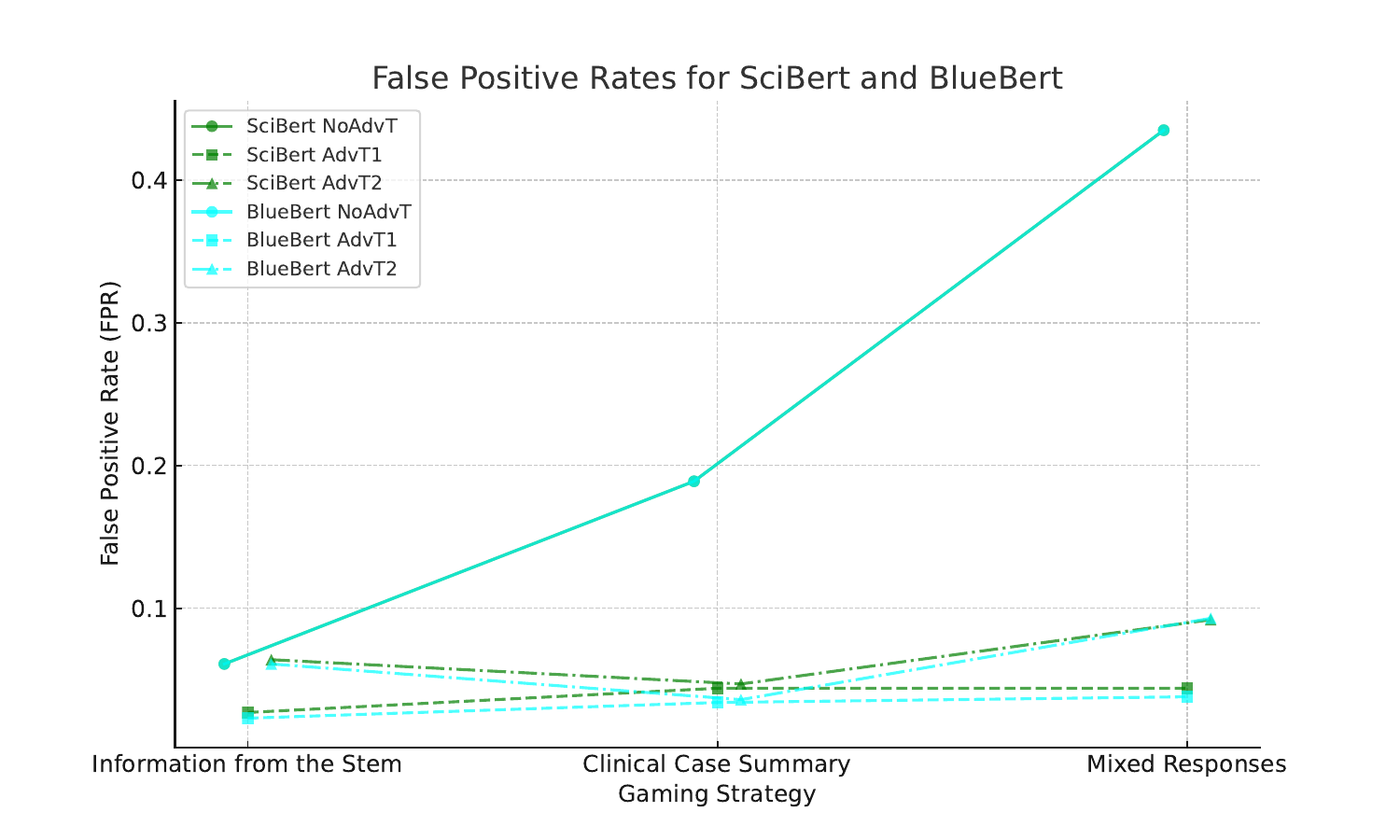} 
        \caption{FPR Across Different Models - Part 2}
        \label{fig:fpr_plots}
    \end{minipage}
\end{figure*}


\begin{table*}
\footnotesize
\centering
\begin{tabular}{|c|c|c|c|}
\hline
\textbf{Gaming Strategy} & \textbf{FPR Before Adv Training} & \textbf{FPR Adv Training \#1}  & \textbf{FPR Adv Training \#2} \\ \hline
Information from the Stem    &  .061           &  .017              & .067    \\ \hline
Clinical Case Summary    &   .189          &    .036               &   .04      \\ \hline
Mixed Responses & .435    &  .041     & .067 \\ \hline
\end{tabular}
\caption{False positive rates for the gaming responses before and after adversarial training}
\label{tab:detector_table}

\end{table*}

To enhance the resilience of the ACTA system against gaming responses, two adversarial training experiments were undertaken to investigate (i) whether adversarial training based on all three types of gaming responses improves system robustness to these types of responses and (ii) whether adversarial training based on two types of gaming responses improves robustness to a third type of responses. The general workflow is demonstrated in Figure  \ref{fig_workflow}.
The first experiment entailed the inclusion of 70\% of the simulated responses from each strategy into the training dataset (together with the authentic responses), with the remaining 30\% of both artificial and authentic responses allocated to the test set. The objective of the second experiment was to assess the capacity of data derived from specific strategies to bolster the model’s defenses against gaming strategies that were not identified during the training phase. This was achieved through the implementation of a 3-fold cross-validation method, where the model was trained on data from two gaming strategies and tested on the third. This approach enabled the evaluation of the model’s enhanced ability to recognize unknown examples through exposure to known gaming adversarial examples.

To enhance the results and evaluate the efficacy of various models, we employed five different models for response embeddings to predict whether a response was related to gaming: Clinical-BERT \cite{huang2019clinicalbert}, Bio-BERT \cite{lee2020biobert}, Sci-BERT \cite{beltagy2019scibert}, and Blue-BERT \cite{peng2019transfer}. These models are pretrained on medical domain datasets, leveraging their specialized knowledge to aid in training the system. They were fine-tuned with the adversarial data in a 70/30 split (for experiment 1) and fine-tuned with two gaming strategies, and tested on the third one 
for experiment 2, as detailed above. These fine-tuned models are then used to classify responses as correct or incorrect. The embeddings generated by these models were then combined with the ACTA model using a majority vote method and ridge regression to determine if there was an improvement. 

\begin{figure}[htbp]
    \centering
    \includegraphics[width=\columnwidth]{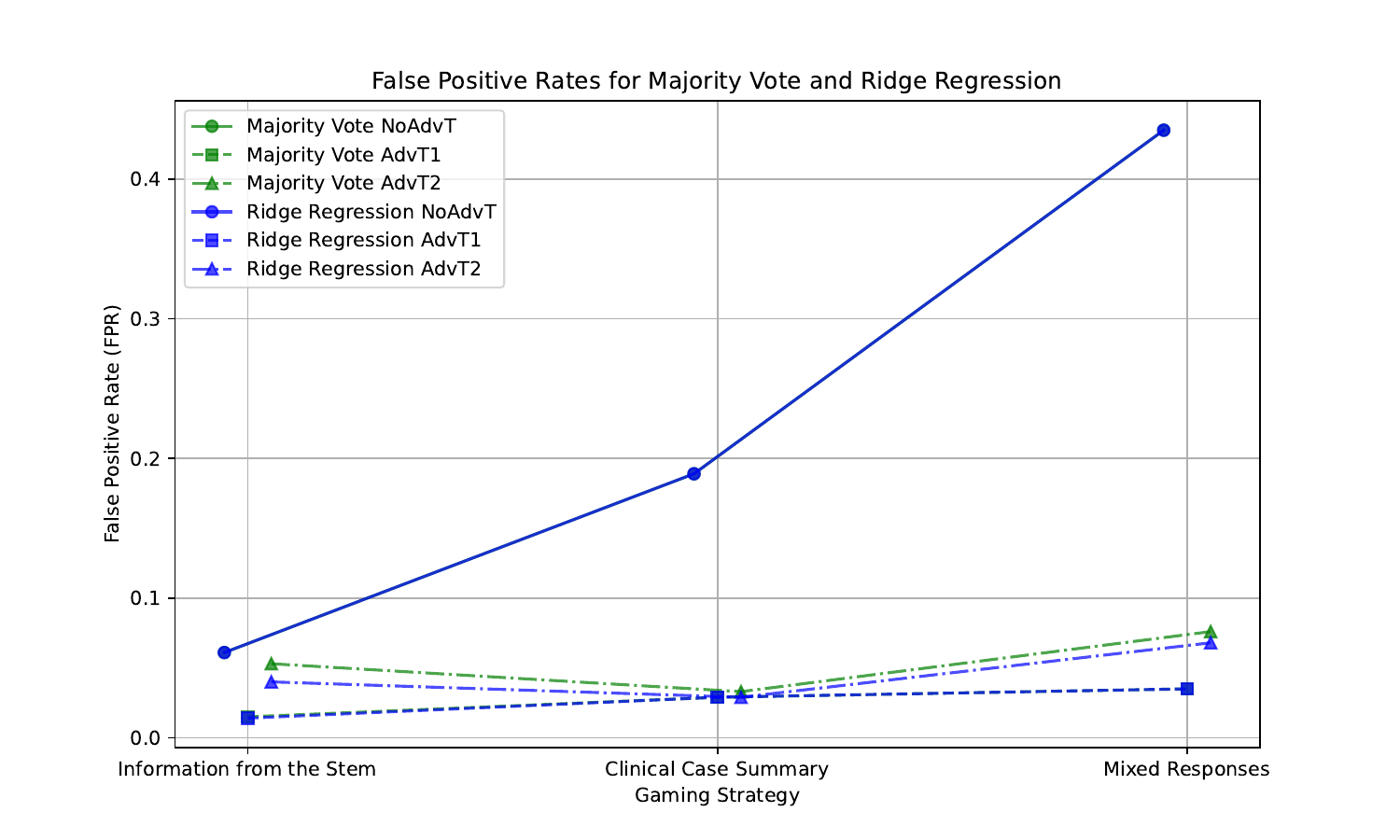}
    \caption{False Positive Rates for Majority Vote and Ridge Regression}
    \label{fig:fpr_combined}
\end{figure}

\subsection{Post Adversarial Training Results}


\renewcommand{\arraystretch}{1.5} 

\begin{table*}[htbp]
    \centering
    \footnotesize
    \renewcommand{\arraystretch}{1.2}
    \begin{tabular}{lccc|ccc}
        \toprule
        \multirow{2}{*}{Gaming strategy} & \multicolumn{3}{c}{Acta Model} & \multicolumn{3}{c}{ClinicalBert} \\
        \cmidrule(lr){2-4} \cmidrule(lr){5-7}
        & FPR (NoAdvT) & FPR (AdvT1) & FPR (AdvT2) & FPR (NoAdvT) & FPR (AdvT1) & FPR (AdvT2) \\
        \midrule
        Information from stem & \textbf{0.061} & \textbf{0.017} & 0.067 & \textbf{0.061} & \textbf{0.023} & 0.060 \\
        Clinical case summary & 0.189 & 0.036 & \textbf{0.040} & 0.189 & 0.033 & \textbf{0.040} \\
        Mixed responses & 0.435 & 0.041 & 0.067 & 0.435 & 0.038 & 0.080 \\
        \midrule
        \multirow{2}{*}{Gaming strategy} & \multicolumn{3}{c}{BioBert} & \multicolumn{3}{c}{SciBert} \\
        \cmidrule(lr){2-4} \cmidrule(lr){5-7}
        & FPR (NoAdvT) & FPR (AdvT1) & FPR (AdvT2) & FPR (NoAdvT) & FPR (AdvT1) & FPR (AdvT2) \\
        \midrule
        Information from stem & \textbf{0.061} & \textbf{0.023} & 0.064 & \textbf{0.061} & \textbf{0.027} & 0.064 \\
        Clinical case summary & 0.189 & 0.034 & \textbf{0.037} & 0.189 & 0.044 & \textbf{0.047} \\
        Mixed responses & 0.435 & 0.039 & 0.082 & 0.435 & 0.044 & 0.092 \\
        \midrule
        \multirow{2}{*}{Gaming strategy} & \multicolumn{3}{c}{BlueBert} \\
        \cmidrule(lr){2-4}
        & FPR (NoAdvT) & FPR (AdvT1) & FPR (AdvT2) \\
        \midrule
        Information from stem & \textbf{0.061} & \textbf{0.023} & 0.061 \\
        Clinical case summary & 0.189 & 0.034 & \textbf{0.036} \\
        Mixed responses & 0.435 & 0.038 & 0.093 \\
        \bottomrule
    \end{tabular}
    \caption{False positive rates for the gaming responses before and after adversarial training using various models}
    \label{tab:fpr_comparison}
\end{table*}

\begin{table*}[t]
\footnotesize
\resizebox{\textwidth}{!}{%
\begin{tabular}{@{}lccc|ccc@{}}
\cmidrule(l){2-7}
 & \multicolumn{3}{c}{\textbf{Majority Vote Model}} & \multicolumn{3}{c}{\textbf{Ridge Regression}} \\ \midrule
\multicolumn{1}{l|}{\textbf{Gaming strategy}} & \textbf{FPR (NoAdvT)} & \textbf{FPR (AdvT1)} & \multicolumn{1}{c|}{\textbf{FPR (AdvT2)}} & \textbf{FPR (NoAdvT)} & \textbf{FPR (AdvT1)} & \textbf{FPR (AdvT2)} \\ \midrule
\multicolumn{1}{l|}{Information from stem} & 0.061 & \textbf{0.015} & \multicolumn{1}{c|}{0.053} & 0.061 & \textbf{0.014} & 0.040 \\
\multicolumn{1}{l|}{Clinical case summary} & 0.189 & 0.029 & \multicolumn{1}{c|}{\textbf{0.033}} & 0.189 & 0.029 & \textbf{0.029} \\
\multicolumn{1}{l|}{Mixed responses} & 0.435 & 0.035 & \multicolumn{1}{c|}{0.076} & 0.435 & 0.035 & 0.068 \\ \bottomrule
\end{tabular}%
}
\caption{False positive rates for the gaming responses before and after adversarial training using Majority Vote and Ridge Regression models}
\label{tab:results2}
\end{table*}

The results from the experiments described above are shown in Figures \ref{fig:fpr_plots} and \ref{fig:fpr_combined} and Tables \ref{tab:fpr_comparison} and \ref{tab:results2}.
In the first experiment, the model maintained a high F1 score, with substantial reductions in FPRs across various gaming strategies and embedding models. This demonstrates the efficiency of adversarial training in enhancing model accuracy. The FPR results for the gaming strategy ``Information from the Stem" were consistently the lowest across models, indicating that even without adversarial training, this model recognized these responses better than the other two gaming strategies. The post-adversarial training gains for the ``Mixed Responses" strategy are particularly encouraging, suggesting that training on simulated gaming responses is an effective countermeasure against the most successful gaming strategy. This highlights the significant benefits of adversarial training for defending against complex adversarial attacks.

The second experiment also maintained a high F1 score of 0.98 for real responses, while still providing some improvements with gaming detection.  These results suggest that familiarity with known gaming strategies helps the model recognize responses based on unknown gaming strategies, enhancing overall robustness. 
The model’s resilience is significantly bolstered by training with `strong’ gaming examples (high FPR) instead of `weak’ ones. The model’s performance was least effective under strategy 3; however, incorporating this strategy into adversarial training markedly improved model efficiency against strategies 1 and 2. In contrast, training with the relatively weaker strategies 1 and 2 yielded lesser improvements in detecting strategy 3, reducing the FPR from 0.435 to 0.067, which is the smallest FPR for the ACTA model in the second experiment among all the models. This observation highlights the intricate relationship between the effectiveness of gaming strategies and the robustness of model training, suggesting a positive correlation where more sophisticated adversarial training leads to improved  robustness.  
Table \ref{tab:results2}  shows the FPR results after applying the embedding models’ results to Majority Vote and Ridge Regression (``AdvT2"). Ridge regression outperformed the majority vote with the FPRs for both experiments 1 and 2 (Figure \ref{fig:fpr_combined}).  
These findings suggest the effectiveness of these two models compared to considering an individual embedding model.  Similar to the embedding model FPR results, gaming strategy 3 is more challenging to recognize and has a higher FPR than gaming strategies 1 and 2. However, FPR results for all gaming strategies are improved compared to each embedding model’s results.

\subsection{Prompt Engineering Experiments}
Three prompting techniques were employed in this experiment:
\begin{enumerate}[noitemsep]
\item The model was provided with the item questions and the examinee’s response to the question. The model was then asked to score the response, given the question.
\item The model was given the questions along with examples of correct answers for each question. The model was then asked to score the examinee’s response.
\item The model was provided with examples of correct answers only, and then asked to score the examinee’s response.
\end{enumerate}

Using ChatGPT-4, scores using each of these strategies were obtained. Due to resource limitations, 100 samples of each gaming data and real data were used for these experiments.

\begin{table}[t]
\footnotesize
\centering
\begin{tabular}{@{}lccc@{}}
\toprule
\textbf{Gaming strategy} & \textbf{Accuracy} & \textbf{TNR} & \textbf{FPR} \\ \midrule
Information from stem & 0.89 & 0.89 & 0.11 \\
Clinical case summary & 0.97 & 0.97 & 0.03 \\
Mixed responses & 0.99 & 0.99 & 0.01 \\ \bottomrule
\end{tabular}
\caption{ChatGPT results for the gaming responses}
\label{tab:results4}
\end{table}

\subsection{Post Prompt Engineering Results}

Summary results from the prompt engineering experiments are shown in Table \ref{tab:results4}. 
Because it performed best overall, only results from the first prompting strategy, submitting a question and a response and requesting a score, are reported.  

For the experiment with real data, the model maintained high performance, with an accuracy of 0.93, a precision of 0.97, and False Positive Rate (FPR) of 0.06. For the gaming data, the highest accuracy was achieved for the third gaming strategy, ``submit multiple answers", with an accuracy of 0.99, a TNR of 0.99, and the lowest FPR of 0.01. This suggests that in this experiment, ChatGPT-4 was more successful in recognizing and scoring responses generated using the third strategy compared to the adversarial training approaches reported above. The second gaming strategy, ``summarize item vignette", also performed strongly with an accuracy of 0.97 and an FPR of 0.028. The first strategy, ``copy words from the item vignette",  had the lowest performance among the gaming strategies, with an accuracy of 0.89 and an FPR of 0.11. These results underscore the model’s effectiveness in handling various gaming strategies, with notable success in the third strategy, and relative ineffectiveness with responses from the first strategy.

\section{Error Analysis} 

\subsection{Error Analysis For Adversarial Training}

The model’s performance was notably better for gaming strategies it was trained on compared to those it had not encountered during training. This points to a potential overfitting issue, where the model becomes too specialized in detecting known adversarial patterns but may struggle with novel or unseen strategies. Despite the reductions in FPRs, some gaming strategies, particularly Strategy 3 (``Mixed Responses"), remained challenging for the model to detect. This suggests that while adversarial training improves the model’s defenses, it may not fully mitigate all vulnerabilities, especially for more sophisticated or nuanced gaming strategies. The quality and representativeness of the adversarial examples used in training had a significant impact on the model’s performance. Training with ``strong`` adversarial examples (those with high FPRs) led to more substantial improvements in robustness, whereas training with ``weaker" examples provided less benefit. This underscores the importance of carefully selecting adversarial examples that accurately reflect the types of gaming strategies the model might encounter in real-world applications.  The cross-validation experiments demonstrated that while training on multiple gaming strategies can enhance the model’s generalization capabilities, the process is complex and computationally expensive. 

\subsection{Error Analysis For Prompt Engineering}

Upon reviewing the rationales across various datasets, several common patterns emerged that explain why certain responses were predicted wrong. 

\textbf{Summary of the clinical scenario:} Many rationales indicate that a response ``aligns with the intended correct pattern" or ``matches the expected correct response." This suggests that the system recognizes patterns it anticipates, regardless of the response's accuracy. For instance, if a response correctly lists all the symptoms of a disease, the model may consider it correct simply because it aligns with the expected diagnosis related to those symptoms. Some rationales reveal that the presence of specific keywords in the responses triggers the system to mark it as correct, e.g., phrases like ``man, 36, suffers sleepiness, ED, weight gain, hypertension" match key descriptors associated with the correct answers, such as ``sleep study." 

If a response mentions symptoms that suggest a disease, the model may consider it correct, even if the actual cause of the disease differs. An example would be a response stating, ``Man on anti-malaria drugs shows signs of hemolysis," where the correct answer is ``Hemolysis due to G6PD deficiency". In this case, because the hemolysis disease was mentioned in the response, the model scored this response as correct.

\textbf{Utilize mixed responses:}
Here, the rationales often point to specific phrases within the response, indicating that the model matches exact or nearly exact phrases it expects, regardless of whether the combination is logically sound. If a response includes correct elements alongside irrelevant parts that do not negate the correct diagnosis, the model may still consider it correct. For instance, ``Rheumatic fever" might be irrelevant, but it does not invalidate the correct diagnosis of ``systemic sclerosis (scleroderma)." Sometimes, the model assesses the overall picture of the response; if a disease shares similarities with another mentioned in the response, it may still be considered correct. For example, the response ``chronic obstructive pulmonary disease bronchiectasis" might be deemed correct because ``bronchiectasis" was the intended correct answer, and it shares similarities with ``chronic obstructive pulmonary disease bronchiectasis".

\textbf{Randomly sampling words:} This strategy involves the use of random words; in cases where the model erroneously produces a correct score, the sample words are general and provide no specific clues about the disease. In such cases, the model relies on the question and uses the information provided to predict the disease, ultimately considering the response correct, although there was not any correct information in the response. 

\textbf{Real Dataset:}
If a response contains minor misspellings, the model may consider it incorrect, even if it matches the correct response. Conversely, the model may consider a vague term correct if it encompasses the specific diagnoses listed. For example, the response ``heart disease" might be accepted as correct, even if the correct answer is a specific type of heart failure or disease. The rationales sometimes rely on broad medical logic. The model might still consider it correct when a response refers to a general disease category without specifying details or subcategories.  This suggests that the model applies standard medical reasoning but may lack the subtlety needed to distinguish between similar conditions. 
In some cases where the general concept is correct but details are slightly different, the model may still mark the response as wrong despite its correctness. These patterns indicate that the model prioritizes exact matches and penalizes variations, even when the overall concept is correct, highlighting its limitations in understanding nuanced or slightly varied responses.


\section{Discussion}

These results add new evidence related to exploitable vulnerabilities in transformer-based grading systems. Despite being artificially generated approximations of potential gaming behaviors, all three gaming strategies were successful in deceiving the non-adversarial trained system. This aligns with findings from previous research, which also reported that adversarial approaches could compromise the integrity of automated systems, particularly when the system is not specifically trained to recognize such attacks \cite{baldwin2025vulnerability}. 
The first group of adversarial training experiments showed that data augmentation is a promising way to fortify ASAG systems against such attacks. 
The cross-validation experiments also showed that it is beneficial to train on examples across gaming strategies, suggesting a transfer of learning between strategies, which holds the potential to protect against unforeseen gaming tactics that may arise in practice.

The results show that incorporating embedding models into Majority Vote and Ridge Regression significantly reduced the false positive rates (FPR) in experiments, which is in line with findings from research on ensemble learning methods that demonstrate their superiority in reducing error rates \cite{naderalvojoud2023improving}. 
Among the gaming strategies evaluated, strategy 3 proved to be the most challenging to recognize, yielding higher FPRs than strategies 1 and 2. Despite this, the FPR results across all gaming strategies showed improvement when compared to the results of each individual embedding model.  This mirrors findings in previous studies, where certain adversarial strategies consistently posed greater challenges to detection systems.

The experiments demonstrated the first prompting strategy was effective, where the model was given questions and responses to score. With real data, the model showed high accuracy and precision and a low FPR, indicating robust performance in evaluating genuine responses. For gaming data, the best results were seen in the ``submit multiple answers" strategy (consistent with \citet{baldwin2025vulnerability}). The ``summarize item vignette" strategy also performed well; however, the ``copy words from the item vignette" strategy performed relatively poorly.  

In summary, while the non-adversarially trained system was susceptible to gaming, the defense mechanisms explored in this paper showed significant reduction in FPR both when training within strategy and across strategies. As the understanding of possible gaming strategies in the context of medical education matures, future work will include the simulation of new adversarial attacks for ASAG systems that are more closely aligned with human behaviors as well as further experimentation with adversarial training. 
Employing regularization techniques such as dropout, weight decay, and early stopping can limit overfitting, which may improve a model’s generalizability. Furthermore, employing various prompt engineering techniques with LLMs also has the potential to enhance performance.

\section{Limitations and Ethical Considerations}

While this study provides promising directions for improving robustness in ASAG systems, several limitations must be acknowledged. First, the adversarial examples used in our experiments are simulated approximations of gaming strategies, rather than authentic, organically derived examples from real-world test-takers. As such, while the strategies are plausible and their effectiveness in gaming the scoring system was proven, they may not fully reflect the diversity and nuance of actual test-taker behaviors, particularly in high-stakes environments. Furthermore, the experiments were conducted within a single domain and dataset, and the generalizability of the findings to other domains—such as legal education, K-12, or general writing assessment—remains uncertain. Different domains may involve distinct response styles, expectations, learner populations, and gaming behaviors, which could impact the effectiveness of adversarial training strategies. Last but not least, this study explored the effects of these gaming strategies on the ACTA scoring system and on using GPT-4 to score responses via prompt engineering. The extent to which these results generalize to other transformer-based or few-shot scoring systems is an open question.

From an ethical standpoint, adversarial training raises important questions related to fairness, transparency, and trust in AI-based scoring. While improving robustness is a core goal, it is also critical to ensure that ASAG systems do not unfairly penalize legitimate test-taking strategies or linguistic variability, especially among non-native speakers or individuals from underrepresented groups. 

It should also be recognized that research into gaming strategies inherently raises concerns about dual-use. While our intention is to strengthen the integrity of ASAG systems, the publication of methods for generating adversarial responses could inadvertently aid malicious actors. To mitigate this risk, we have intentionally abstracted implementation details and focused on generalizable insights rather than system-specific exploits.

On the positive side, adversarial examples can serve an additional purpose in enhancing explainability. When used in conjunction with feature attribution methods, adversarial perturbations can help identify which aspects of a response most influence model predictions. For example, if minor lexical changes significantly affect scoring, it may indicate an over-reliance on specific keywords or surface features rather than deeper semantic understanding. For example, the error analyses of the prompt engineering approach revealed that the models tend to recognize anticipated patterns as a proxy to accuracy, which is what makes them particularly susceptible to gaming responses that follow the expected pattern of correct answers. These insights are critical for diagnosing model weaknesses, refining scoring rubrics, and improving transparency. In high-stakes assessment, the ability to explain and justify model decisions is essential for fostering user trust and ensuring accountability in automated assessment.

Overall, while adversarial training is a valuable tool for increasing the reliability of ASAG systems, its application must be guided by ethical principles that prioritize fairness, interpretability, and alignment with educational values.


\bibliography{acl_latex}

\appendix

\section{Example Appendix}
\label{sec:appendix}

This is an appendix.

\end{document}